# Using The Feedback of Dynamic Active-Pixel Vision Sensor (Davis) to Prevent Slip in Real Time

Armin Masoumian
Department of Computer Engineering and Mathematics
Universitat Rovira I Virgili
Tarragona, Spain
e-mail: masoumian.armin@gmail.com

Pezhman Kazemi
Departament of Chemical Engineering
Universitat Rovira I Virgili
Tarragona, Spain
e-mail: pezhman.kazemi@urv.cat

Mohammad Chehreghani Montazer
Mechatronics Department
University of Debrecen
Debrecen, Hungary
e-mail: mava.994@gmail.com

Hatem A. Rashwan
Department of Computer Engineering and Mathematics
Universitat Rovira I Virgili
Tarragona, Spain
e-mail: hatem.abdellatif@urv.cat

Domenec Puig Valls
Department of Computer Engineering and Mathematics
Universitat Rovira I Virgili
Tarragona, Spain
e-mail: domenec.puig@urv.cat

*Abstract*—The objective of this paper is to describe an approach to detect the slip and contact force in real-time feedback. In this novel approach DAVIS camera used as a vision tactile sensor due to its fast process speed and high resolution. Two hundred experiments were performed on four objects with different shape, size, weight and material to compare the accuracy and respond of the Baxter robot grippers to avoid slipping. The advanced approach is validated by using a force-sensitive resistor (FSR402). The events captured with DAVIS camera are processed with specific algorithms to provide feedback to the Baxter robot aiding it to detect the slip.

*Keywords-vision sensor; tactile sensor; baxter robot; real-time feedback; vision-based slip detection*

## I. INTRODUCTION

The world has been moving up on robotics very much. Robotics has found its way into almost all the sectors of manufacturing. The main feature of implementing robotics is the repeatability and precision of the job done. The most recent addition to robotics is the use of artificial intelligence which is capable of decision making to increase the efficiency of the given job. On the other hand robots are also faster and precise than human labor. Hence robotics have these qualities which make them useful in any field of work.

The most common robot used is the robotic arm which is capable of lifting and moving objects. The main function of the robotic arm is to identify said objects and move them to the desired location. This is achieved by the use of sensors on the robot. Sensors are devices that interact with the environment and generate electrical signals based on their interaction. The electrical signals are processed by a microcontroller which gives meaningful values as feedback to the robot. Conventional sensors used in a robotic arm are force sensor, tactile sensor and touch sensor. The feedback from these sensors are used to discern any slippage while lifting the object and adjust the holding force of the gripper accordingly. These sensors provide a "sense of touch" for the robot's hand.

Though the sensors are good in detecting slip of the objects they will not be able to apply the minimum force required so that the object does not deform. This will be destructive in the case of fragile objects. The slip is detected only when a shear force is produced and the object slips away. The idea of using a camera to detect the slip will overcome these disadvantages. The input from the camera can also be used in conjunction with deep learning algorithms to benefit from incipient slip detection.

The Baxter robot is an industrial robot built by rethink robotics [1]. It has two arms with an animated face. Since the robot is mainly used in assembly and manufacturing lines, a safety feature has been built into the robot which is capable of sensing humans near-by and restricting its motion. Thus it eliminates the need to be in a caged working environment unlike other robots. The Baxter achieves this by having multiple sensors in its hand and arms, while the motors actuate the joints through springs. The springs provide a cushioning effect and make it less hazardous as it is able to reduce the force before collision. The Baxter robot is capable of learning complicated movements in a unique way. The robot's hand can be moved physically by a person in the desired way and the robot is capable of memorizing the motion and replicating it. There is a single mounted camera on Baxter last joint which can be used for teaching Baxter to do tasks [2].

For this project, a transparent gripper needs to get the events of the contacted area, therefore, a new gripper designed with acrylic and transparent silicone.

The DAVIS camera is a unique type of camera, the method in which it acquires inputs sets it apart from the traditional cameras. The DAVIS camera does not capture frames rather it captures events (independent pixels that have intensity changes) in an asynchronous manner. Therefore, pixels that have intensity changes are only captured and the output of the camera is not a sequence of images but a stream of asynchronous events.

The DAVIS camera is an entirely different way of approach in the field of vision sensors. It is inspired by the function of the human eye and tries to mimic it.

Knowing how this approach benefits functioning for eyes, researchers are developing machine-vision frameworks in which every pixel modifies its own inspecting because of changes in the measure of incident light it gets. What's expected to actualize this plan is electronic hardware that can track the amplitudes of every pixel constantly and record changes of just those pixels that move in light level by some little recommended sum. This method of capturing data is called level-crossing sampling.

Each pixel is attached to a level-crossing detector and a separate exposure-measurement circuit. The amplitudes of each pixel are measured and checked in a reference to a previously measured threshold value or compared with the previous signal, if below the previous signal the new level is recorded and used as reference. So, each pixel changes its sampling rate depending on the level-crossing. With this course of action, if the measure of light reaching a given pixel changes rapidly, that pixel is refreshed frequently. In the case of nothing changes, the pixel quits getting what might simply end up being repetitive data and goes sit outs until the point that changes begin to happen again.

All neuromorphic chips sold by iniVation utilize the AER (Address Event Representation) convention to transmit events in a nonconcurrent way [3]. The AER convention is a straightforward convention utilizing a variable number of lines to transmit information, and two lines (REQ and ACK) to synchronize the information between the sender and the recipient utilizing a four-stage handshake. (This is likewise called a packaged nonconcurrent convention). The ACK and REQ lines are dynamic low.

In addition, the DAVIS camera can be used to detect the deformation of the object during grasping by the Baxter grippers [4].

This camera has a high-speed computer vision because it works mostly with event-based, which has low latency, high temporal resolution, and very high dynamic range [5].

## II. Experimental Setup

Traditional vision sensors procure the visual data as an arrangement of pictures, recorded at discrete intervals in time. Visual data gets time quantized at a fixed rate which has no connection to the elements active in the scene. Besides, each recorded frame passes on the data from all pixels, paying little respect to whether this data or a piece of it, has changed since the last frame had been procured. This obtaining technique constrains the fleeting determination, possibly missing imperative data, and prompts repetition in the recorded picture information, all the while blowing up information rate and volume. With this in mind, the decision was taken to shift to neuromorphic vision systems. Since the event camera only gets pixels that change they have advantages over traditional cameras. They do not process redundant information (only captures pixels that change) hence they save a lot of memory, do not waste computational energy on redundant pixels, do not suffer from latency as they do not have to wait for unnecessary frames and has a very high temporal resolution, very high dynamic range and low power and bandwidth requirements.

DAVIS240C camera has been used for detecting the slip of the object in a millisecond and then sending the feedback of the DAVIS camera to the gripper of the Baxter robot. DAVIS camera has been mounted on behind the grippers with a specific angle, which based on the brightness of the light, will detect the touching area of the object and gripper. Therefore, gripper must be transparent so the camera can see the object through the gripper, and gripper must be made of soft tissue materials so that by deformation of the gripper camera will detect whether or not slip is happen. Because of those reasons, transparent silicone has been chosen for this task because silicone is a soft tissue material and can get deformation based on the applying force to the object.

Transparency and thickness of the silicone grippers are quite important in this task because much transparency can cause a lot of noise in the background of the object. If the thickness of the silicone is big, the contact area will not be clear in the DAVIS camera and if the thickness is too thin, then it will not capture the deformation the silicone and object can slip easily through the grippers.

To prove that force is changing in Baxter grippers, force sensitive resistor has been set and coded in Raspberry PI 2B to validate the value of the force. The FSR402 is a Force Sensitive Resistor which is a robust polymer thick film (PTF) device which shows a decrease in resistance by increasing the force applied to the sensor surface. FSR has been developed to use in electronic devices, which needs to be controlled by human touch such as automotive electronics, medical systems. FSR402 will show the force in a range of 0 to 1024, although this number can be changed based on a resistor that has been set for this sensor. This resistance depends on the amount of pressure that will be applied on the sensor which is more force will cost less resistance and less force will make high resistance and if there is no pressure on the sensing area, the resistance will be bigger than $1M\Omega$. This sensor can sense applied force anywhere in the range of 100g-10kg [6].

The Raspberry PI is a small single-board computer, which has different types and models. This raspberry PI that has been used in this project is 2B. There are two USB port 2 and 1 Ethernet socket in this model. For communicating between Raspberry PI and Baxter robot, an Ethernet socket has been used.

Robot Operation System, known as ROS, is an open-source mixture of middleware and operating systems for robots. This meta-operating system provides the services that expect from a robot operating system such as hardware abstraction, low-level device control, implementation of commonly used functionality, message passing between processes, and package management. One of the most advantages of ROS is that it can provide a huge amount of tools for configuring the setup for the robot, start programming, testing and analyzing the results debugging the problems and many more possibilities. In addition, ROS has the availability to use many libraries that implement useful robot functionality, with a focus on mobility, manipulation, and perception [7].

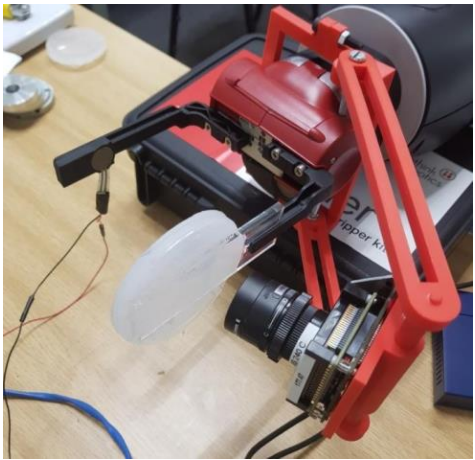

Figure 1. Assembly Setup on Baxter Gripper.

Socket programming is the most fundamental method of programming for communicating with the network. Sockets will create a port between operating systems to communicate with each other. In a simple way, the socket is the easiest way to connect 2 pieces of software together which called point to point connection.

C Address-Event Representation, known as cAER, is a software framework for event processing, targeted mainly towards embedded systems, also with visualization capabilities that make cAER suitable for desktop and research use. Low-power embedded systems are a natural match for low-power neuromorphic sensors and processors. cAER is written in C/C++, and does not require a GUI to function and can also be completely controlled via network.

For running the DAVIS camera with cAER program, there is a davis-config file with XML format, which needs to be configured and set to receive the data from the camera. In this file, some sections such as the background noise filter, the output module for receiving the outputs and the OpenCV for visualizing the framework of the camera has been added. In this part, the Noise filter module had two purposes of using, which was counting the number of polarity events at a certain time and also capturing the pixel locations of the polarities.

Analogue image processing will analyze and manipulate the hard copy of the image. However, digital image processing will analyze and manipulate the image in a computer with image processing software.

The noise filter is used to filter out noises in the background of the frame that is focused upon. The noise filter works by using the concept of hot pixels. Hot pixels can be considered as an abnormal behavior of pixels. An object with a specific size would produce at least more than two or three pixels depending upon the distance. If the object is too far away it is entirely a different story. Any change of an object very far away would be represented as a single pixel.

For learning which pixels are hot pixels or not, the two parameters, which is explained above, is taken as an input form the user. The time period in microseconds to accumulate events for learning hot pixels and the number of events needed in a learning time period to consider a pixel hot. The figures below show the difference between the noise filter and without noise filter.

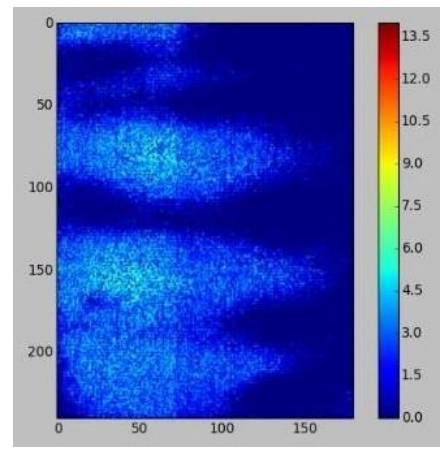

Figure 2. Contact area without noise-filtering.

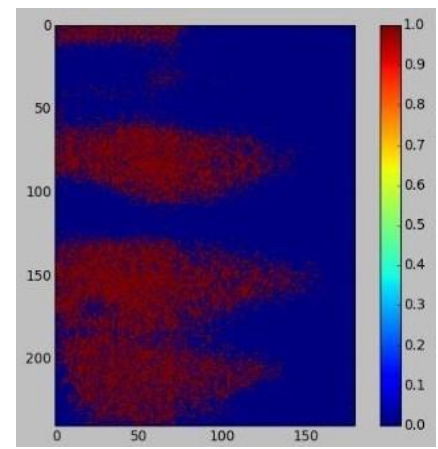

Figure 3. Contact area with noise-filtering.

The contact area is the area of the object that is touching the silicone in the gripper. Finding out the contact area helps to determine the slip of the object efficiently. The contact area is determined by capturing the events from the camera when the gripper is closing in on the objects. Objects coming in contact with silicone will produce negative polarity events at the point of contact because they block the incoming light to the camera.

Based on the results from image processing, if the camera detects the slip happening, then, the gripper position will be reduced to prevent slip. This was the feedback between the camera and the Baxter robot and is done with socket programming. However, the important part is gripper must know how much needed to be closed or opened to stop slipping. Because if the gripper position is being reduced too much, then it will deform or even break the object. For finding the best values for the gripper position in each period, the PID controller has been programmed for this task, which will be explained in the next section.

There are two parameters of the gripper that can be used to control the force applied to the object in- between the grippers namely, holding force and gripper position. The holding force determines how much resistive force the robot applies to counter change from its current gripper position. The gripper position specifies the distance between the grippers in terms of percentage with hundred a being fully open and zero as being fully closed. The absolute value in terms of length depends upon the configuration of the gripper, which can be changed manually by removing the gripper and repositioning them.

### III. PID Control

The PID controller is employed in this research as a means of feedback and to control the force of the grippers. The gripper position parameter is used as a means of controlling the force. The PID controller works by receiving an error, which is a deviation from a specified set point and adjusts the control outputs in order to drive the error to zero. The three values namely, proportional, integral and derivative determine how quickly and accurately the error is driven to zero. The error in this setup would the number of positive polarity events captured in the contact area while the program is running. Also, the error has to be updated after each timeframe, called the sample time. The sample time chosen for this setup is one millisecond. The PID is updated regularly after each sample time specified above. The error term has to pass through certain filtering explained in the previous section in order to eliminate noises. Since the contact area matrix was normalized the polarity events captured have equal weighting and the corresponding response. The setpoint, in this case, is zero as we want no positive polarity events, as we intend to eliminate slipping.

Three combinations of PD, Pi and PID have been tested for this project and based on the experiments, PI was the best option to use for this task because the proportional controller would be the initial response that reduces error and acts aggressively and might even overshoot exceeding the force required to control slip. This is where the derivative term helps to reduce the overshoot. The derivative term keeps track of the rate of change of error with respect to the time. Depending upon the sign and magnitude of the term the derivative controller reduces the output accordingly to prevent overshoot. The proportional term acts when the error is detected, but it goes to zero when there is no error. This will oscillate the system, in other words, the gripper will reduce the command position when slip is detected and when there is no slip the gripper will go back to the initial position facilitating slip again putting the system in an endless cycle and this is not good for this task. However, the integral term keeps track of all the previous errors that occurred and has an integral sum over time. This term eventually adds up to large term and takes over to hold the gripper in place according to the previously occurred errors. Note that the integral term doesn't drop to zero when no error is detected, unlike the proportional term. This ensures that the controller always holds the gripper in the required position to stop the slip of the object based on previous errors. Hence, the PI controller has been chosen for this project.

### IV. Result

For each object, two parameters of the gripper position and force sensor value have been discussed. The first graph represents the gripper positions of the gripper and the second graph represents the force sensor values. Based on the 200 experiments, cube shape objects have more accurate results than spherical objects because the contact area is bigger. Because of the vibration of robot arms, there are a little fluctuation in force sensor values.

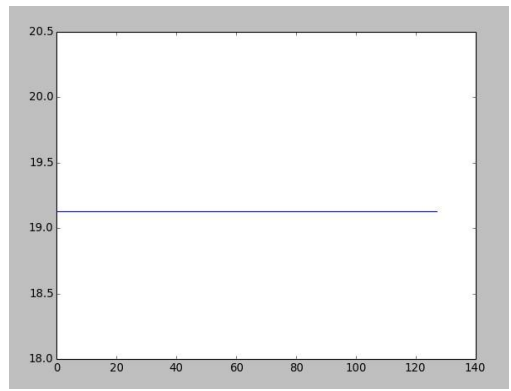

Figure 4. Gripper position values for plastic box 110 gr.

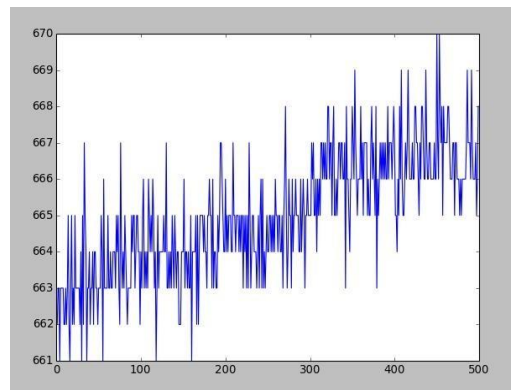

Figure 5. Force sensor values for plastic box 110 gr.

Fig. 4 represents the gripper position values for the light object, therefore there is no slip. By increasing the weight,

the object will slip more as it is shown in Fig. 6. The reason of dropping down the force values in the fig. 7 is because the object is slipping, the contact area between the object and force sensor will be lost and because of that the force values will be decreased, but after gripper will reduce the position and slip will be stopped, then again contact area will be stables and force values will be increased slightly.

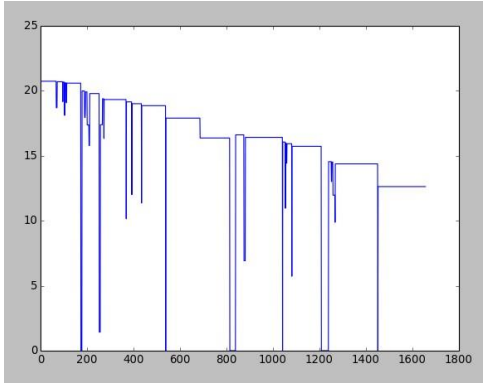

Figure 6. Gripper position values for plastic box 1360 gr.

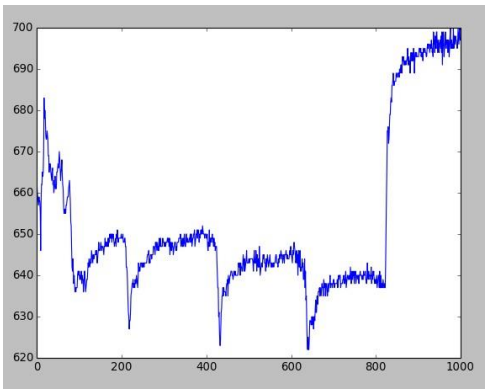

Figure 7. Force sensor values for plastic box 1360 gr.

Sometimes there is a difference between number of force values and gripper position values, because the times, which slip is not happening, the results will be same and gripper will not count the values, but force sensor will capture all values till one cycle period time will be down, which in this experiment all cycle were 20 seconds.

## V. CONCLUSION

In this paper, a novel method approach was proposed to detect the slip and contact force in real-time feedback. DAVIS camera was used as a vision tactile sensor for fast process and high resolution. Socket programming was also used to get close-loop in real-time feedback. The advanced approach is validated by using a force-sensitive resistor (FSR402). The events captured with the DAVIS camera were processed with algorithms to provide feedback to the Baxter robot aiding it to detect slip. All devices are synchronized to avoid any delays. PID controller was used for stabilizing the system to get more accurate results.